\documentclass{article}

\usepackage[utf8]{inputenc} % allow utf-8 input
\usepackage[T1]{fontenc}    % use 8-bit T1 fonts
\usepackage{hyperref}       % hyperlinks
\usepackage{url}            % simple URL typesetting
\usepackage{booktabs}       % professional-quality tables
\usepackage{amsfonts}       % blackboard math symbols
\usepackage{nicefrac}       % compact symbols for 1/2, etc.
\usepackage{microtype}      % microtypography
\usepackage[pdftex]{graphicx}

\usepackage[preprint]{neurips_2020}

\title{Predicting Nanorobot Shapes via Generative Models}

\author{%
  Emma Benjaminson\\
  Mechanical Eng.\\
  Carnegie Mellon University\\
  Pittsburgh, PA 15213\\
  \texttt{ebenjami@andrew.cmu.edu}\\
    \And 
  Rebecca E. Taylor\\
  Mechanical Eng.\\
  Biomedical Eng.\\
  Electrical \& Computer Eng.\\
  Carnegie Mellon University\\
  Pittsburgh, PA 15213\\
  \texttt{bex@andrew.cmu.edu}\\
  \And
  Matthew Travers\\
  Robotics Institute\\
  Carnegie Mellon University\\
  Pittsburgh, PA 15213\\
  \texttt{mtravers@andrew.cmu.edu}\\

}

\begin{document}

\maketitle

\begin{abstract}
 The field of DNA nanotechnology has made it possible to assemble, with high yields, different structures that have actionable properties. For example, researchers have created components that can be actuated, used to sense (e.g., changes in pH), or to store and release loads. An exciting next step is to combine these components into multifunctional nanorobots that could, potentially, perform complex tasks like swimming to a target location in the human body, detect an adverse reaction and then release a drug load to stop it. However, as we start to assemble more complex nanorobots, the yield of the desired nanorobot begins to decrease as the number of possible component combinations increases. Therefore, the ultimate goal of this work is to develop a predictive model to maximize yield. However, training predictive models typically requires a large dataset. For the nanorobots we are interested in assembling, this will be difficult to collect. This is because high-fidelity data, which allows us to exactly characterize the shape and size of individual structures, is very time-consuming to collect, whereas low-fidelity data is readily available but only captures bulk statistics for different processes. Therefore, this work combines low- and high-fidelity data to train a generative model using a two-step process. We first use a relatively small, high-fidelity dataset to train a generative model. Then at run time, the model takes low-fidelity data and uses it to approximate the high-fidelity content. We do this by biasing the model towards samples with specific properties as measured by low-fidelity data. In this work we bias our distribution towards a desired node degree of a graphical model that we take as a surrogate representation of the nanorobots that this work will ultimately focus on. We have not yet accumulated a high-fidelity dataset of nanorobots, so we leverage the MolGAN architecture [1] and the QM9 small molecule dataset [2-3] to demonstrate our approach. 
\end{abstract}
 
%  Therefore, the ultimate goal of this work is to learn to control the assembly process to maximize yield. We can accelerate the learning process with a predictive model trained on experimental data. 

\section{Introduction}

The field of DNA nanotechnology has generated a rich array of nanoscale designs for applications such as moving through fluids [4], sensing changes in pH [5-6] and storing drug loads [7]. We hypothesize that we could combine components with individual capabilities into multifunctional nanorobots to perform complex tasks, potentially, like swimming through the human body to a target site, detecting an adverse reaction, and then releasing a large drug load to stop the reaction from continuing. We could build these nanorobots by combining multiple species of functional components in solution under specific manufacturing conditions, such as temperature and ion concentration. However, research suggests that as we build more complex nanorobots, the yields will decrease [8]. The engineering challenge here is to learn to optimize the manufacturing conditions to maximize yield, and we can accelerate this learning process with a predictive model. 

However, training predictive models typically requires a large dataset, which will be difficult to collect in this application. This is because the dimensions of nanoscale objects are below the diffraction limit of light, so we must use time-consuming techniques like superresolution microscopy, atomic force microscopy (AFM) or transmission electron microscopy (TEM) to image the nanorobots directly. The alternative to using these high-fidelity methods is to use low-fidelity characterization techniques like spectrophotometry. Spectrophotometry uses light to characterize sample composition and can be used to detect the number of unhybridized DNA strands in solution, which can be correlated to how many connected neighbors a component has. If we represent the structure of a nanorobot as a graph, we may be able to relate the spectrophotometry data to the average node degree of that graph. Therefore, while spectrophotometry data is faster to collect than AFM or TEM, it can only tell us how many connections there are between components in a nanorobot, but not the types of components present in the nanorobot, nor its overall topology. Note that spectrophotometry techniques are not the only low-fidelity tools available to us - we could also use gel electrophoresis or fluorescence microscopy, among others. In future, we will conduct experiments to understand what characteristics we can reliably observe about our nanorobots from low-fidelity data, and we will use those metrics to bias our model's output using the same approach as the one described here. 

\begin{figure}
    \centering
    \fbox{\includegraphics[width=0.9\textwidth]{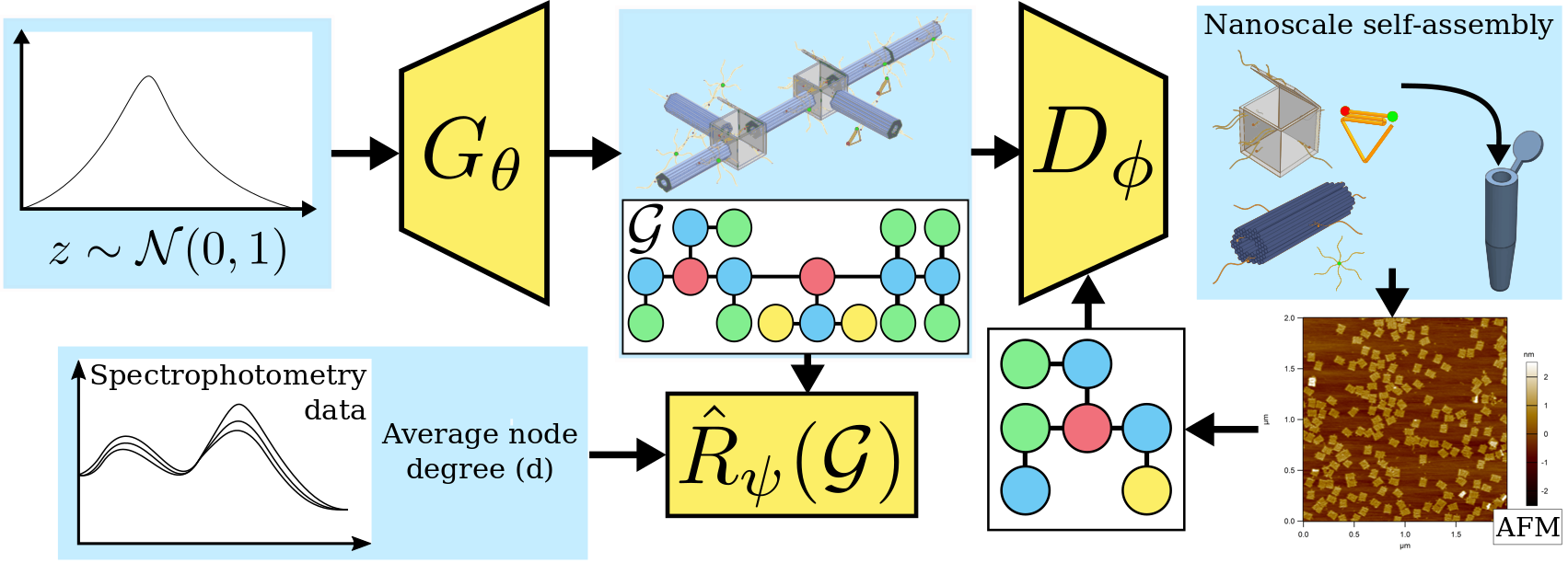}}
    \caption{An overview of our two-step training approach that uses high-fidelity (AFM) data and low-fidelity (spectrophotometry data) to train a GAN with an additional reward network. (Credit to Ying Liu for AFM image.)}
    \label{gan_structure}
\end{figure}

We propose to bridge the gap between time-consuming imaging techniques and fast, yet low-fidelity characterizations by combining them to train a generative model in a two-step process. We support the belief that generative models are an appropriate choice because they can learn to represent a distribution and make novel predictions by biasing the learned distribution towards desired characteristics. Our two-step training process, as shown in Figure~\ref{gan_structure}, will first train the model on a relatively small dataset of high-fidelity data. (Based on work in [1], we estimate we will only need a few thousand example nanorobots.) Once the model has learned to approximate the distribution of nanorobot shapes, the second step will be to collect only low-fidelity data for subsequent experiments. This data will then be used to bias the distribution towards nanorobots with a desired average node degree.  

% Our GAN consists of the generator, $G_{\theta}$, the discriminator, $D_{\phi}$ and the reward network, $\hat{R}_{\psi}$. The generator pulls random samples and returns graphs that represent nanoscale superstructures. The discriminator guides the generator to return graphs, $\mathcal{G}$, that look similar to those drawn from AFM data from the underlying nanorobot manufacturing process that we are studying. The reward network biases the output of the generator to return graphs with a particular average node degree as estimated from spectrophotometry data. 

While there are many examples of generative models that incorporate a mechanism for adding bias to their outputs [1, 9-14], we selected the MolGAN architecture presented by De Cao and Kipf for our preliminary experiments [1]. The MolGAN framework was designed to learn a distribution of small molecules (maximum 9 atoms) from the QM9 dataset [2-3], represented as graphs. MolGAN linearly combines the typical loss function for the generator with a reinforcement learning (RL) objective in order to bias the generator to output novel molecules that meet certain criteria, such as solubility or druglikeliness (a term coined in [15]) [1]. 

We present our preliminary implementation of a generative model, as inspired by MolGAN, using a novel reward objective. We compare our model's performance to the original MolGAN using the same QM9 dataset employed by De Cao and Kipf [1]. Note that we use the QM9 dataset [2-3] as a surrogate for our DNA nanorobots because we do not yet have a dataset of nanorobot structures. Since we intend to represent our nanorobots as graphs, the QM9 dataset is a good surrogate because it also uses a graph-based representation. This allows us to test our novel reward objective, which is designed to bias the generator towards returning graphs with a desired average node degree. In this work we explore two aspects of the model's performance that will be of particular importance moving forwards: (1) how well we can bias our model's output and (2) how diverse the generated samples are when we are biasing the model. 

\section{Key Concepts from Prior Work}

In this section, we present key concepts from the literature that we used in our implementation. First, we define a graph as a set of nodes, $V$, connected by a set of edges, $E$ which together represent a graph, $G = (V, E)$. Following De Cao and Kipf, we can encode information about the nodes and edges in one-hot tensors. For example, the nodes are described by a feature matrix, $X$, which has size $N \times T$, where $N$ is the number of nodes in the graph, and $T$ is the number of different types of nodes. Similarly, the edges can be described by an adjacency tensor, $A$, of size $N \times N \times B$, where $B$ is the number of different edge types [1]. 

%, including the Wasserstein GAN with gradient penalty (WGAN-GP) [cite], and the relational graph convolutional network (R-GCN) used by the discriminator and reward networks [cite]. We will provide a brief explanation of how the Deep Deterministic Policy Gradient (DDPG) algorithm [cite] is used to help the generator learn from both the discriminator and the reward network, and conclude by presenting our novel reward objective

\subsection{Wasserstein Generative Adversarial Networks with Gradient Penalty (WGAN-GP)}

% The generative adversarial network (GAN) is designed to implicitly represent the distribution of a dataset [16]. The original version of a GAN is made up of two components: a generator, $G_{\theta}$, which produces novel samples, and a discriminator, $D_{\phi}$, which learns to determine if a given sample is real (i.e. from the training set) or fake (i.e. produced by the generator). 
The MolGAN architecture uses a variant on the original GAN [16] that minimizes the Wasserstein-1 distance instead of the Kullbeck-Leibler divergence; this variant is called a WGAN [17]. The advantage of the WGAN is that it is less prone to mode collapse than the original GAN because the Wasserstein-1 distance function is much less susceptible to the vanishing gradient problem [17]. 

The Wasserstein-1 distance function is intractable to compute, so it is approximated using the Kantorovich-Rubinstein duality [17-18]. This approximation uses a K-Lipschitz function that must, by definition, be bounded. Gulrajani et al. suggested using a gradient penalty to apply this constraint, giving rise to the WGAN-GP [19]. This WGAN-GP model is used in MolGAN, and allows us to define a new loss function for the discriminator as follows [1]:
\begin{equation}\label{grad_penalty}
    L\big(\textbf{x}^{(i)}, G_{\theta}(\textbf{z}^{(i)}); \phi \big) = -D_{\phi}(\textbf{x}^{(i)}) + D_{\phi}(G_{\theta}(\textbf{z}^{(i)})) + \alpha\bigg(||\nabla_{\hat{\textbf{x}}^{(i)}} D_{\phi}(\hat{\textbf{x}}^{(i)})|| - 1\bigg)^2
\end{equation}
Where $D_{\phi}$ is the discriminator, $G_{\theta}$ is the generator, and $\hat{\textbf{x}}^{(i)}$ is a linear combination of real and fake samples, $\hat{\textbf{x}}^{(i)} = \epsilon \mathbf{x}^{(i)} + (1-\epsilon)G_{\theta}(\mathbf{z}^{(i)})$. The random seed is $\textbf{z}^{(i)} \sim \mathcal{N}(0,1)$ and $\epsilon \sim \mathcal{U}(0,1)$ [1]. We set $\alpha = 10$ following Gulrajani et al.'s recommendation [19]. 

\subsection{Relational Graph Convolutional Networks (R-GCN)}

We use graph convolution in both the discriminator and reward networks as a means of finding salient graph features that help each network to categorize graphs as real or fake (in the case of the discriminator) and to attach the correct score to each graph for meeting the design criteria (in the case of the reward function). In MolGAN, De Cao and Kipf use a variant of the Relational Graph Convolutional Network (R-GCN) by Schlichtkrull et al. [20-21] to obtain the following convolution function [1]: 
\begin{equation}
    H_i^{(l+1)} = \tanh \bigg( \sum_{j=1}^N \sum_{y=1}^Y \frac{\tilde{A}_{ijy}}{|\mathcal{N}_i|} f_y^{(l)}(H_j^{(l)}, x_i) + f_s^{(l)}(H_i^{(l)}, x_i) \bigg)
\end{equation}
$H_i^{(l+1)}$ represents the output of a graph convolution layer. The first term sums up the graph convolution for every node, $x_i$, in the previous layer, $H_j^{(l)}$, with its neighbors in $\tilde{A}$, a modified version of the adjacency matrix. Dividing by $|\mathcal{N}_i|$ (the number of neighbors of the $i$-th node) normalizes the output. The function $f_y^{(l)}$ is from a family of affine functions for every edge type, and it performs the graph convolution step in the R-GCN. Similarly, $f_s$ is a function that adds self-loops to the output of each layer [20]. 

We must aggregate the output of the final layer of the R-GCN into a scalar value for the discriminator and the reward networks. We use a soft-attention mechanism to obtain some "attention glimpse," or scalar, as follows [1]: 
\begin{equation}
    H_{\mathcal{G}} = \tanh \bigg( \sum_{v \in \mathcal{V}} \sigma(i(H_v^{(l)}, x_v)) \odot \tanh(j(H_v^{(l)}, x_v)) \bigg)
\end{equation}
Where $H_{\mathcal{G}}$ is the scalar output of the soft-attention mechanism as applied to the entire graph, $\mathcal{G}$. Two different multilayer perceptrons are represented by $i$ and $j$, $\sigma$ is the sigmoid activation function, and $\odot$ is element-wise multiplication [1]. 

\subsection{Deep Deterministic Policy Gradient (DDPG) and the Generator Loss Function}
% The MolGAN architecture is unique because it also incorporates a reward network that helps guide the output of the generator. The reward network is trained against a reward function as described in the next section [1]. 

De Cao and Kipf framed the graph generation process as a one-step RL episode where the actor (i.e. the generator) must choose an action (i.e. an output graph) from a high-dimensional action space. Presented this way, the deep deterministic policy gradient (DDPG) algorithm [22] is an appropriate method of learning to optimize the actor's policy because it is capable of operating in high-dimensional spaces with neural networks. De Cao and Kipf cast the generator as the policy of the MolGAN agent, and the reward function serves as the critic in this paradigm. The authors use a streamlined version of the DDPG approach, neglecting the use of experience replay and target networks, because we are working with i.i.d. data in this case [1]. 

% \footnote{Note that although the number of possible types of atoms in a graph, and the graph size, are limited, the number of possible output graphs is very large. This fact makes the action space high-dimensional.}

Ultimately, we can write the loss function for the generator as follows [1]: 
\begin{equation}
    L(\mathbf{z}^{(i)}; \theta) = \lambda \big(-D_{\phi}(G_{\theta}(z))\big) + (1 - \lambda)\big(-\hat{R}_{\psi}(G_{\theta}(z))\big)
\end{equation}
Where $\hat{R}_{\psi}$ is the output of the reward network, and $\lambda$ is a hyperparameter for tuning the trade-off between the WGAN component and the RL component of the generator's loss [1]. 

\section{Results}

% In this section we present our novel reward function, details about our implementation of a generative model as inspired by MolGAN, and the experiments that we conducted to characterize its performance.

\subsection{Novel Reward Objective}

The MolGAN architecture uses a neural network to approximate a specific reward function provided by the user. This function approximator provides a gradient that the generator can follow to maximize reward. The reward network must minimize the mean squared error between its predictions and the actual outputs from the reward function ($R$) for both real and fake data [1]: 
\begin{equation}
    L\big(G_{\theta}(\textbf{z}^{(i)}), \mathbf{x}^{(i)}; \psi\big) = \bigg(\hat{R}_{\psi}(G_{\theta}(\textbf{z}^{(i)}) - R(G_{\theta}(\textbf{z}^{(i)}) \bigg)^2 + \bigg(\hat{R}_{\psi}(\mathbf{x}^{(i)}) - R(\mathbf{x}^{(i)}) \bigg)^2
\end{equation}
In this paper we created a novel reward function that computes the average node degree for every graph in a set. The function assigns a score in the range [0, 1] based on how closely the average node degree for a set of graphs matches the desired value, similar to the MolGAN approach [1]. 

\subsection{Implementation}
\begin{table}[b]
  \caption{Effect of reward function objective ($d$) on output bias and sample diversity. The bolded values in the MolGAN data indicate the maximum values for each score.}
  \label{exp1_results}
  \centering
  \begin{tabular}{lllll}
    \toprule
    \multicolumn{5}{l}{\textbf{Original MolGAN [1]}}                   \\
    \cmidrule(r){1-5}
    Objective     & \% Unique  & Druglikeliness & Synthesizability & Solubility \\
    \midrule
    Druglikeliness & 2.2  & \textbf{0.62} & 0.59 & 0.53    \\
    Synthesizability & 2.1 & 0.53 & \textbf{0.95} & 0.68 \\
    Solubility & 2.0 & 0.44 & 0.22 & \textbf{0.89} \\
    All Simultaneously & 2.3 & 0.51 & 0.82 & 0.69 \\
    \midrule
    \midrule
    \multicolumn{5}{l}{\textbf{Our Results}}                   \\
    \cmidrule(r){1-3}
    Objective & \% Unique & \multicolumn{3}{l}{Average $d$} \\
    \cmidrule(r){1-3}
    $d = 1$ & $0.1$ & $1.1$ (average over 3 trials)\\
    $d = 2$ & $0.8$ & $2.0$ \\
    $d = 4$ & $0.7$ & $4.0$ \\
    $d = 6$ & $0.7$ & $6.0$ (average over 3 trials)\\
    Baseline $(\lambda = 1.0)$ & $12.8$ & $1.4$ \\
    \cmidrule(r){1-3}
  \end{tabular}
\end{table}
We conducted a series of experiments with our novel reward function to investigate two key questions: (1) how well did our novel reward function bias the output of our generator towards the desired average node degree? And (2) how diverse was our output sample set for varying values of $\lambda$?

In accordance with the test conditions established by De Cao and Kipf, in each experiment we trained\footnote{We wrote our implementation in TensorFlow v2.3.0 [23]. We chose to use RMSProp [24] ($\alpha = $1e-3, $\rho = 0.9$, momentum$= 0$) instead of Adam [25] as our optimizer based on the recommendation by Arjovsky et al. which observed that using momentum-based optimizers tended to perform worse in training a WGAN than those that optimized without momentum [17].} the appropriate version of our model for 300 epochs on a 5K random sampling from the QM9 dataset [2-3] (we also performed a validation run on an additional 1664 samples). We pretrained the model for the first 150 epochs (i.e. $\lambda = 1.0$) and then for the second half of the training process, we set $\lambda$ to the appropriate value. In the results shown below, we computed statistics for 6400 sampled graphs, after repeating for 5 trials and averaging, unless otherwise stated.

\subsection{Effect of Reward Function on Output Bias}

First, we varied the desired average node degree, $d$, in our reward function and trained our model to bias its outputs to meet this objective. For all of these experiments, we set $\lambda = 0$ for the second half of training. This is following the precedent set by De Cao and Kipf, who found that $\lambda = 0$ maximized the generator's scores from the reward function. We trained the model to target values of $d = \{1, 2, 4, 6\}$ and report the resulting average node degree for each value of $d$ in Table~\ref{exp1_results}. We also report the percentage of unique (non-isomorphic) graphs. Our results are compared to those from De Cao and Kipf in [1] and [26]. Figure~\ref{results} shows samples of graphs generated by the models. 

\subsection{Effect of $\lambda$ on Output Diversity}

Second, we investigated the effect that $\lambda$ had on the diversity of samples from the generator. We measured this as the percentage of graphs from a set that were unique (n = 6400 for each of 5 trials unless otherwise stated) after training the model with values of $\lambda = \{1.0, 0.75, 0.5, 0.25, 0.05, 0.0\}$. We set $d = 2$ for all trials and also report the resulting average node degree. The results are shown in Table~\ref{exp2_results}, and we compare them to De Cao and Kipf's findings in [1],\footnote{De Cao and Kipf computed this as the ratio of all non-isomorphic graphs to all valid graphs. Here we consider all graphs as valid so we calculated the percentage of \textit{all} non-isomorphic graphs in a set of 6400 samples from the original dataset.} including their model's performance with respect to solubility, which they were attempting to optimize with their reward function. We also compare to an additional dataset describing sample diversity and quantitative estimate of druglikeliness (QED) [15] provided in [26] by De Cao.
\begin{table}[t]
  \caption{Effect of $\lambda$ on output bias and sample diversity.}
  \label{exp2_results}
  \centering
  \begin{tabular}{lllllll}
    \toprule
     & \multicolumn{2}{l}{\textbf{MolGAN [1]}}& \multicolumn{2}{l}{\textbf{MolGAN} [26]} & \multicolumn{2}{l}{\textbf{Our Results}}      \\       
    \cmidrule(r){1-7}
    $\lambda$     & \% Unique  & Solubility & \% Unique & QED & \% Unique & Average $d$ \\
    \midrule
    0.0 & 2.3 & 0.86 & 3.16 & 0.61 & $0.8$ & $2.0$ \\
    0.05 & 2.5 & 0.67 & - & - & $2.3$ & $2.0$ \\
    0.25 & 1.9 & 0.65 & 10.16 & 0.61 & $6.3$ & $2.0$ \\
    0.5 & 1.8 & 0.48 & 31.29 & 0.56 & $8.9$ & $2.0$  \\
    0.75 & 2.5 & 0.57 & 64.35 & 0.51 & $9.8$ & $2.1$ \\
    1.0 & 2.5 & 0.54 & 63.91 & 0.50 & $12.8$ & $1.4$ \\
    \bottomrule
  \end{tabular}
\end{table}
\section{Discussion}
In our first experiment, we observed that our novel reward function was able to bias the average node degree of the sample graphs obtained after training. The average $d$ values reported agreed closely with the desired values. The added bias also greatly reduced the percentage of graphs that were non-isomorphic, from $12.8\%$ to less than $1\%$ for most values of $d$. These results suggest that although the model can be biased towards graphs with specific characteristics, it comes at a cost of greatly reduced output sample diversity. 

\begin{figure}
    \centering
    \fbox{\includegraphics[width=0.9\textwidth]{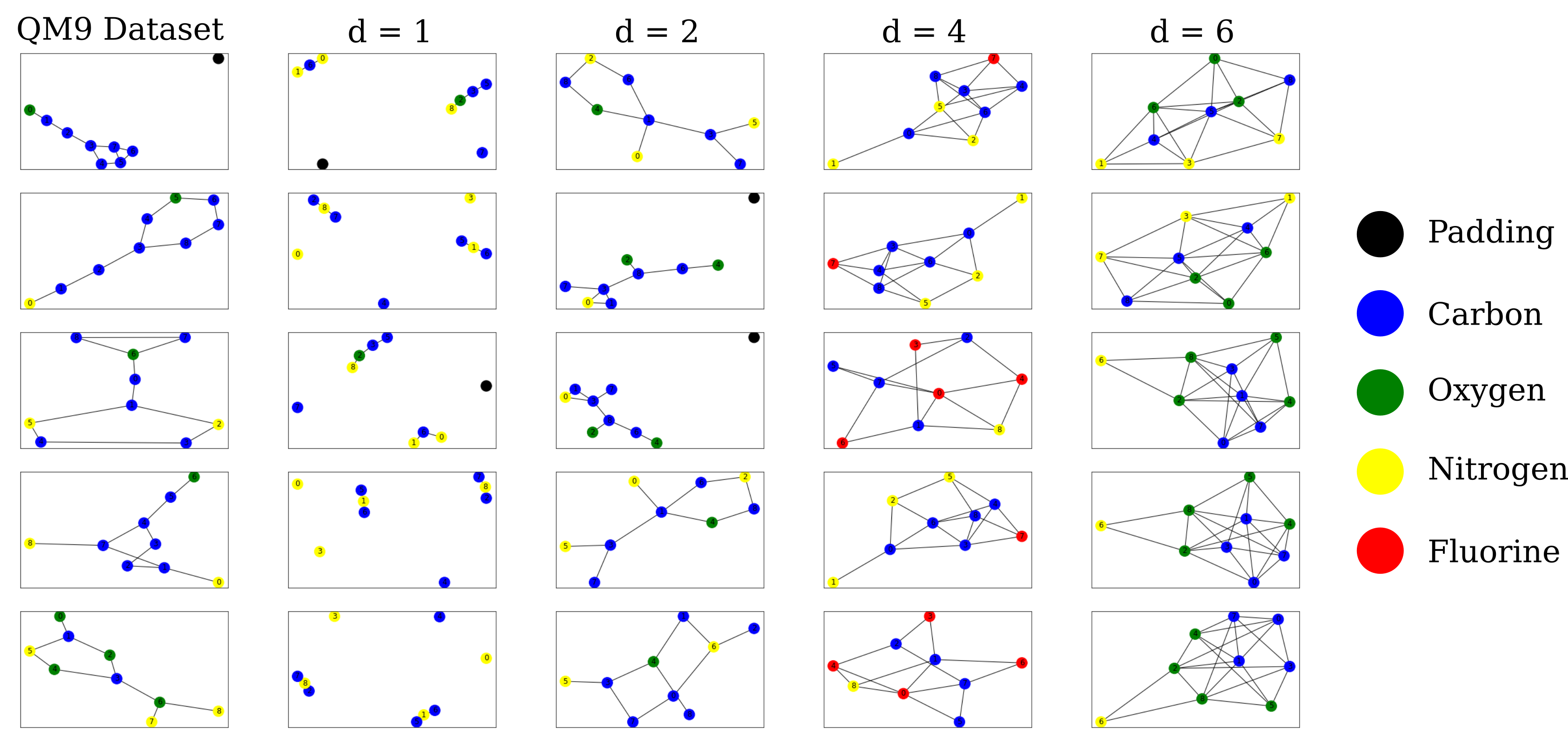}}
    \caption{Sample graphs for every value of $d$ tested in this work, compared to some sample graphs from the QM9 dataset. (All graphs were generated using NetworkX [27].)}
    \label{results}
\end{figure}

In reviewing Figure~\ref{results}, it is clear that as $d$ increased, the sample graphs also showed increasing numbers of connections between nodes. Interestingly, the model seems to have also been biased towards using more fluorine atoms in the $d = 4$ test. The $d = 2$ samples seem the most closely related to the QM9 dataset, and show indications that the model had learned to form benzene-like rings. 

The variation of our results with added bias disagreed with the trend that De Cao and Kipf found in their original paper [1]. Specifically, they found that bias did not significantly affect the uniqueness values as compared to tests conducted without added bias [1]. Perhaps our objective - a specific average node degree - is a more stringent constraint than De Cao and Kipf's objectives of druglikeliness, synthesizability and solubility. 

In our second experiment, we found that as we decreased $\lambda$, we saw a corresponding decrease in the percentage of unique graphs in our sample set. This suggests that as we pursued the RL objective more heavily (i.e. as $\lambda$ decreased), the generator experienced mode collapse and only output a few unique graphs that met the desired average node degree criteria. This result disagrees with the findings presented by De Cao and Kipf in [1], but agrees with findings in [26]. In [1], the $\%$ unique score did not vary with $\lambda$, but in [26] $\%$ uniqueness decreased along with $\lambda$. Interestingly, we also observed that as we weighted the RL objective more heavily, the average node degree values converged on $2.0$. 

\section{Conclusions}
In this preliminary investigation, we used a novel reward function to bias the output of a generative model. The results presented suggest that this approach is capable of generating novel samples that are similar to the training dataset and also possess desired characteristics as dictated by the reward function. We intend to pursue this approach further to predict the shape of nanorobots formed with DNA nanotechnology components, using only low-fidelity data. 

However, our data also showed reduced sample diversity with added bias, which could potentially limit our ability to accurately predict all the possible nanorobots we would see \textit{in vitro}. In future work we will investigate alternative approaches to biasing the output of a GAN so that we can preserve our ability to output graphs with specific characteristics while expanding the diversity of the output. 

\section*{Broader Impact}
This paper is intended to benefit other researchers who wish to apply generative models to their own research by demonstrating a new approach. It is always possible that our model and our approach to adding bias could be used to tune the output graphs for discriminatory or dangerous purposes. There is limited risk of bias in the QM9 dataset [2-3] what we used, because it is a collection of small organic molecules (up to 9 heavy atoms), and so should not contain any human-based discriminatory bias.

% Given the nascent stage of this research direction, however, we do not believe that this work puts anyone at a disadvantage at this time. Failure of the model only leads to non-sensical or limited generated graphs, which is unlikely to have any detrimental impact on researchers or society as a whole at this time.

\begin{ack}
The authors have no competing financial interests to disclose. This material is based on work supported by the National Science Foundation under Grant No. 1739308 and by the National Science Foundation Graduate Research Fellowship Program under Grant No. DGE1745016, awarded to Emma Benjaminson. Any opinions, findings, and conclusions or recommendations expressed in this material are those of the authors and do not necessarily reflect the views of the National Science Foundation. Thank you to members of the Microsystems and Mechanobiology Lab and the Biorobotics Lab for many valuable discussions. Thank you also to Nicola De Cao and Faruk Ahmed for their personal correspondence in conjunction with this work.
\end{ack}

\section*{References}

% References follow the acknowledgments. Use unnumbered first-level heading for
% the references. Any choice of citation style is acceptable as long as you are
% consistent. It is permissible to reduce the font size to \verb+small+ (9 point)
% when listing the references.
% {\bf Note that the Reference section does not count towards the eight pages of content that are allowed.}
% \medskip

\small

[1] De Cao, N. \& Kipf, T. (2018) MolGAN: An implicit generative model for small molecular graphs. \textit{arXiv Prepr. arXiv:1805.11973}.

[2] Ruddigkeit, L., van Deursen, R., Blum, L. C. \& Reymond, J.-L. (2012) Enumeration of 166 billion organic small molecules in the chemical universe database GDB-17. \textit{Journal of Chemical Information and Modeling} \textbf{52}(11):2864–2875.

[3] Ramakrishnan, R., Dral, P. O., Rupp, M. \& von Lilienfeld, O. A. (2014) Quantum chemistry structures and properties of 134 kilo molecules. \textit{Scientific Data} \textbf{1}, 140022. 

[4] Dreyfus, R., Baudry, J., Roper, M. L., Fermigier, M., Stone, H. A. \& Bibette, J. (2005) Microscopic artificial swimmers. \textit{Nature}, \textbf{437}(6):2–5.

[5] Narayanaswamy, N., Chakraborty, K., Saminathan, A., Zeichner, E., Leung, K. H., Devany, J. \& Krishnan, Y. (2019) A pH-correctable, DNA-based fluorescentreporter for organellar calcium. \textit{Nature Methods}, \textbf{16}(1):95–102.

[6] Liu, Y., Kumar, S. \& Taylor, R. E. (2018) Mix-and-match nanobiosensor design: Logical and spatial programming of biosensors using self-assembled DNA nanostructures. \textit{Wiley Interdisciplinary Reviews:  Nanomedicine and Nanobiotechnology}, pp. e1518.

[7] Douglas, S. M., Bachelet, I. \& Church, G. M. (2012) A logic-gated nanorobot for targeted transport of molecular payloads. \textit{Science}, \textbf{335}(6070):831–834.

[8]  Tikhomirov, G., Petersen, P. \& Qian, L. (2017)  Fractal assembly of micrometre-scale DNA origami arrays with arbitrary patterns. \textit{Nature}, \textbf{552}(7683):67–71.

[9] Radford, A., Metz, L. \& Chintala, S. (2016) Unsupervised representation learning with deep convolutional generative adversarial networks. \textit{4th International Conference on Learning Representations, ICLR 2016 - Conference Track Proceedings}.

[10] Zhu*, J., Park*, T., Isola, P. \& Efros, A.A. (2017) Unpaired Image-to-Image Translation using Cycle-Consistent Adversarial Networks. \textit{IEEE International Conference on Computer Vision (ICCV)}. (* indicates equal contributions) 

[11] Bojchevski, A., Shchur, O., Zügner, D. \& Günnemann, S. (2018) NetGAN: Generating Graphs via Random Walks. \textit{35th International Conference on Machine Learning, ICML 2018}. \textbf{2}:973–988.

[12] Mirza, M. \& Osindero, S. (2014) Conditional Generative Adversarial Nets. \textit{arXiv Prepr. arXiv:1411.1784}.

[13] Kang, S. \& Cho, K. (2019) Conditional Molecular Design with Deep Generative Models. \textit{Journal of Chemical Information and Modeling} \textbf{59}(1):43–52.

[14] Guimaraes, G. L., Sanchez-Lengeling, B., Outeiral, C., Farias, P. L. C. \& Aspuru-Guzik, A. (2017) Objective-Reinforced Generative Adversarial Networks (ORGAN) for Sequence Generation Models. \textit{arXiv Prepr. arXiv:1705.10843}.

[15] Bickerton, G. R., Paolini, G. V., Besnard, J., Muresan, S. \& Hopkins, A. L. (2012) Quantifying the chemical beauty of drugs. \textit{Nature Chemistry}, \textbf{4}(2):90.

[16] Goodfellow, I.J., Pouget-Abadie, J., Mirza, M., Xu, B., Warde-Farley, D., Ozair, S., Courville, A. \& Bengio, Y. (2014) Generative Adversarial Networks. \textit{Proceedings of the 27th International Conference on Neural Information Processing Systems} \textbf{2}:2672-2680.

[17] Arjovsky, M., Chintala, S. \& Bottou, L. (2017) Wasserstein Generative Adversarial Networks. \textit{Proceedings of the 34th International Conference on Machine Learning, PMLR} \textbf{70}:214-223.

[18] Villani, C. (2009) Optimal Transport: Old and New. \textit{Grundlehren der mathematischen Wissenschaften.} Springer, Berlin.

[19] Gulrajani, I.,  Ahmed, F., Arjovsky, M., Dumoulin, V. \& Courville, A. (2017) Improved Training of Wasserstein GANs. \textit{Proceedings of the 31st International Conference on Neural Information Processing Systems}, pp.\ 5769–5779.

[20] Schlichtkrull, M., Kipf, T. N., Bloem, P., van den Berg, R., Titov, I. \& Welling, M. (2018) Modeling Relational Data with Graph Convolutional Networks. Gangemi A. et al. (eds) \textit{The Semantic Web. ESWC 2018. Lecture Notes in Computer Science}, vol 10843. Springer, Cham. https://doi.org/10.1007/978-3-319-93417-4\_38

[21] Kipf, T. N. \& Welling, M. (2016) Semi-Supervised Classification with Graph Convolutional Networks. \textit{5th International Conference on Learning Representations - Conference Track Proceedings}.

[22] Lillicrap, T. P., Hunt, J. J., Pritzel, A., Heess, N., Erez, T., Tassa, Y., Silver, D. \& Wierstra, D. (2015) Continuous control with deep reinforcement learning.\textit{ arXiv preprint
arXiv:1509.02971}.

[23] Martin, A., Barham, P., Chen, J., Chen, Z., Davis, A., Dean, J., Devin, M., Ghemawat, S., Irving, G., Isard, M., Kudlur, M., Levenberg, J., Monga, R., Moore, S., Murray, D. G., Steiner, B., Tucker, P., Vasudevan, V., Warden, P., Wicke, M., Yu, Y. \& Zheng, X. (2016) TensorFlow: A system for large-scale machine learning. 
\textit{12th USENIX Symposium on Operating Systems Design and Implementation (OSDI 16)}, USENIX Association, pp. 265-283

[24] Tieleman, T. \& Hinton, G. (2012) Lecture 6.5—RmsProp: Divide the gradient by a running average of its recent magnitude. \textit{COURSERA: Neural Networks for Machine Learning}.

[25] Kingma, D. P. \& Ba, J. (2014) Adam: A method for stochastic optimization. \textit{CoRR}, abs/1412.6980.

[26] De Cao, N. (2019) Deep Generative Models for Molecular Graphs. Presented at \textit{Machine Learning for Physics and the Physics of Learning 2019, Workshop I: From Passive to Active: Generative and Reinforcement Learning with Physics}, hosted by the Institute for Pure \& Applied Mathematics, UCLA. 

[27] Hagberg, A. A., Schult, D. A. \& Swart, P. J. (2008) Exploring network structure, dynamics, and function using NetworkX. \textit{Proceedings of the 7th Python in Science Conference (SciPy2008)}, Gäel Varoquaux, Travis Vaught, and Jarrod Millman (Eds), (Pasadena, CA USA), pp. 11–15.

\end{document}